\pdfoutput=1

\documentclass[11pt, twocolumn]{article}

\usepackage[]{ACL2023}

\usepackage{times}
\usepackage{latexsym}
\usepackage{hyperref}
\usepackage{graphicx}
\usepackage{verbatim}
\usepackage{booktabs}
\usepackage{float}

\usepackage[T1]{fontenc}

\usepackage[utf8]{inputenc}

\usepackage{microtype}

\usepackage{inconsolata}

\usepackage{multicol} 

\usepackage{inconsolata}
\usepackage{listings}
\usepackage{color}
\usepackage{algorithm} 
\usepackage{algpseudocode} 
\usepackage{graphicx} 
\usepackage{float}    
\usepackage{pifont}
\usepackage{booktabs} 
\usepackage{tabularx} 
\usepackage{geometry} 
\usepackage{amsmath} 

\usepackage{times}
\usepackage{microtype}
\usepackage{epsfig}
\usepackage{caption}
\usepackage{float}
\usepackage{placeins}
\usepackage{color, colortbl}
\usepackage{stfloats}
\usepackage{enumitem}
\usepackage{tabularx}
\usepackage{xstring}
\usepackage{multirow}
\usepackage{xspace}
\usepackage{url}
\usepackage{subcaption}
\usepackage{xcolor}
\usepackage[hang,flushmargin]{footmisc}

\definecolor{mylightgray}{gray}{0.9}

\definecolor{codegreen}{rgb}{0,0.6,0}
\definecolor{codegray}{rgb}{0.5,0.5,0.5}
\definecolor{codepurple}{rgb}{0.58,0,0.82}
\definecolor{backcolour}{rgb}{0.95,0.95,0.92}
\definecolor{mypurple}{RGB}{200,192,248}
\definecolor{mypurpledeep}{RGB}{142,126,240}
\definecolor{mygreen}{RGB}{117,170,156}
\definecolor{myyellow}{RGB}{255,192,0}
\definecolor{myblue}{RGB}{57,143,255}
\definecolor{mygrey}{RGB}{231,230,230}
\definecolor{codey}{RGB}{220,220,170}
\definecolor{coder}{RGB}{206,145,120}
\definecolor{codeb}{RGB}{156,220,254}
\definecolor{codenum}{RGB}{204,204,204}

\DeclareSymbolFont{extraup}{U}{zavm}{m}{n}
\DeclareMathSymbol{\varheart}{\mathalpha}{extraup}{86}
\DeclareMathSymbol{\vardiamond}{\mathalpha}{extraup}{87}

\lstdefinestyle{mystyle}{
    backgroundcolor=\color{backcolour},   
    commentstyle=\color{codegreen},
    keywordstyle=\color{magenta},
    numberstyle=\tiny\color{codegray},
    stringstyle=\color{codepurple},
    basicstyle=\footnotesize,
    breakatwhitespace=false,         
    breaklines=true,                 
    captionpos=b,                    
    keepspaces=true,                 
    numbers=left,                    
    numbersep=5pt,                  
    showspaces=false,                
    showstringspaces=false,
    showtabs=false,                  
    tabsize=2
}

\newcommand{\thankswithlink}[1]{%
  \begingroup
  \renewcommand\thefootnote{}
  \footnote{%
    \noindent
    \fontsize{9.6}{10}\selectfont 
    \href{#1}{#1}%
  }%
  \addtocounter{footnote}{-1}
  \endgroup
}

\newcommand{\modelname}{BioClinical ModernBERT\xspace}

\title{\modelname: A State-of-the-Art Long-Context Encoder for Biomedical and Clinical NLP}

\author{
    Thomas Sounack$^{1}$
  \\\And
   Joshua Davis$^{1,2}$
   \\\And
   Brigitte Durieux$^{1,3}$
  \\\AND
  Antoine Chaffin$^{4}$
  \\\And
  Tom J. Pollard$^{5}$
  \\\And
  Eric Lehman$^{6}$
  \\\And
  Alistair E. W. Johnson$^{7}$
   \\\ANDtwo
  Matthew McDermott $^{8}$
  \\\And
   Tristan Naumann$^{9}$
  \\\And
    Charlotta Lindvall$^{1,8}$
}

\affil{
   $^1$ Dana-Farber Cancer Institute \hspace{0.5cm}
   $^2$ Albany Medical College \hspace{0.5cm}
   $^3$ McGill University \\\vspace{1mm}
   $^4$ LightOn \hspace{0.5cm}
   $^5$ Massachusetts Institute of Technology \hspace{0.5cm}
   $^6$ OpenEvidence \\\vspace{1mm}
   $^7$ Microsoft \hspace{0.5cm}
   $^8$ Harvard Medical School \hspace{0.5cm} 
   $^9$ Microsoft Research
 
}

\corremail{thomas\_sounack@dfci.harvard.edu}

\begin{document}

\maketitle

\thankswithlink{https://github.com/lindvalllab/BioClinical-ModernBERT}

\begin{abstract}

Encoder-based transformer models are central to biomedical and clinical Natural Language Processing (NLP), as their bidirectional self-attention makes them well-suited for efficiently extracting structured information from unstructured text through discriminative tasks. However, encoders have seen slower development compared to decoder models, leading to limited domain adaptation in biomedical and clinical settings. We introduce \modelname, a domain-adapted encoder that builds on the recent ModernBERT release, incorporating long-context processing and substantial improvements in speed and performance for biomedical and clinical NLP. \modelname is developed through continued pre-training on the largest biomedical and clinical corpus to date, with over 53.5 billion tokens, and addresses a key limitation of prior clinical encoders by leveraging 20 datasets from diverse institutions, domains, and geographic regions, rather than relying on data from a single source. It outperforms existing biomedical and clinical encoders on four downstream tasks spanning a broad range of use cases. We release both base (150M parameters) and large (396M parameters) versions of \modelname, along with training checkpoints to support further research.

\end{abstract}

\section{Introduction}

As clinical information databases have become more comprehensive, there has been a surge in the application of Natural Language Processing (NLP) methods to support downstream healthcare tasks, ranging from clinical decision support and care delivery analyses to patient cohort selection for clinical trials~\citep{choTaskSpecificTransformerBasedLanguage2024, abdel-jaberReviewDeepLearning2022}. This trend has coincided with the rapid evolution of deep learning models, particularly transformer-based architectures, which have shown strong performance across both generative and discriminative NLP tasks~\citep{minRecentAdvancesNatural2024}.

Encoder-based architectures, which process entire input sequences using bidirectional attention, have traditionally been preferred for non-generative tasks such as classification and span extraction, due to their ability to capture rich contextual representations. In contrast, decoder-based models rely on autoregressive decoding, generating one token at a time while attending only to previously generated tokens. This structure makes them well-suited for open-ended text generation~\citep{nielsenEncoderVsDecoder2025}. Although encoder-based approaches were once dominant, recent research has increasingly centered on decoder-based models, and encoder-based architectures have received comparatively less interest~\citep{shoolSystematicReviewLarge2025}. This has led decoders to be used for non-generative tasks where encoders would be better suited, using workarounds like structured output generation~\citep{gengJSONSchemaBenchRigorousBenchmark2025}. The community's preference for decoder-based models may at least in part be attributed to their capacity to support longer input lengths and have better generalization as well as zero-shot capabilities; advantages largely driven by greater engineering investment and scaling efforts. Nonetheless, encoders are still necessary in many applications and are used in conjunction with LLMs, such as in Retrieval-Augmented Generation (RAG) frameworks~\citep{lewisRetrievalAugmentedGenerationKnowledgeIntensive2020}.

The release of ModernBERT~\citep{modernbert} represents a significant advancement in encoder architectures, introducing a longer context window, improved efficiency, an expanded vocabulary, and a substantially larger volume of training data. These enhancements are particularly impactful in the clinical domain. With a context window of up to 8,192 tokens, extendable via rotary positional embeddings (RoPE)~\citep{suRoFormerEnhancedTransformer2023}, ModernBERT enables the processing of entire clinical notes and documents in a single pass, eliminating the need for fragmentation into smaller chunks. The increased vocabulary size of 50,368, compared to BERT's 30,000, supports the learning of more precise token embeddings, which is especially beneficial for capturing the diversity and complexity of clinical and biomedical terminology.

In this work, we perform a two-step continued pretraining of the ModernBERT models on biomedical and clinical corpora, offering both base and large versions, and publicly release the trained models along with their training checkpoints. We demonstrate significant improvements, achieving state-of-the-art performance across multiple benchmarks compared to existing encoders, while offering support for long-context inputs and maintaining the highest overall efficiency across various input distributions. Our approach leverages the largest training dataset ever used for a biomedical or clinical encoder and incorporates a diverse range of clinical text sources to supplement the traditionally relied-upon MIMIC data~\citep{johnsonMIMICIIIFreelyAccessible2016, johnsonMIMICIVFreelyAccessible2023}. Notably, while MIMIC replaces protected health information (PHI) with generic de-identification tags, limiting the model’s ability to learn meaningful representations of entity-specific context, several of the additional datasets we include use realistic surrogate identifiers which allows \modelname to learn more natural representations around these entities.
\section{Related Work}

\paragraph{General domain encoders} Although decoder-only models have gained significant popularity, encoder-only architectures remain widely used due to their favorable trade-off between performance and efficiency, especially in Information Retrieval (IR) and RAG pipelines. Their low inference cost and ability to process documents efficiently make them attractive for large-scale applications. While many existing pipelines still rely on older models like BERT, which are limited by short context windows, outdated training corpora, and less efficient design choices~\citep{wangGLUEMultiTaskBenchmark2018, xiaoCPackPackedResources2024}, recent efforts such as MosaicBERT~\citep{portesMosaicBERTBidirectionalEncoder2024}, CrammingBERT~\citep{geipingCrammingTrainingLanguage2022}, and GTE-en-MLM~\citep{liGeneralTextEmbeddings2023a} have introduced updated architectures and longer context support. However, these models have largely focused on retrieval tasks or efficient training rather than broad downstream performance. ModernBERT~\citep{modernbert} addresses many of these limitations by scaling up training data to two trillion tokens, introducing improved architectural choices, and offering competitive performance across a wider range of tasks.

\paragraph{Biomedical and clinical encoders} Recent advancements in NLP have led to the development of domain-adapted transformer models that significantly enhance the extraction and interpretation of clinical text. Among these, BioBERT~\citep{biobert} and Clinical BERT~\citep{alsentzerPubliclyAvailableClinical2019} have emerged as pivotal tools in biomedical and clinical informatics, respectively. BioBERT extends the original BERT architecture through pretraining on large-scale biomedical literature, enabling more accurate semantic understanding of domain-specific language. Clinical BERT builds upon this by incorporating additional pretraining on clinical notes from the MIMIC-III dataset~\citep{johnsonMIMICIIIFreelyAccessible2016}, allowing it to better capture the nuances of clinical documentation. BioMed-RoBERTa~\citep{gururanganDontStopPretraining2020}, relying on the RoBERTa architecture~\citep{liuRoBERTaRobustlyOptimized2019}, takes a different approach by training on Semantic Scholar~\citep{kinneySemanticScholarOpen2025} as its source of biomedical text. These models have outperformed general-domain encoders in tasks such as concept extraction, temporal relation identification, and outcome prediction, making them essential components in the clinical NLP pipeline.

\paragraph{Long context} Long-context understanding is especially important in the clinical NLP domain, as clinical documentation tends to be highly variable in structure, terminology, and length, often exceeding the standard 512-token input limit of many encoder-based models~\citep{ruleLengthRedundancyOutpatient2021}. The ability to process longer sequences enables models to capture and integrate relevant information dispersed across entire clinical notes or even multiple documents. This extended context facilitates the identification of patterns and relationships that shorter-context models often miss. As a result, long-context encoders have demonstrated improved performance on a range of clinical NLP tasks, such as phenotyping, cohort selection, and medical entity recognition. This is exemplified by~\citet{liClinicalLongformerClinicalBigBirdTransformers2022}, who introduced Clinical-Longformer and Clinical-BigBird, currently the only publicly available clinical models capable of handling extended sequence lengths.
\section{Methods}

\begin{table*}[hb!]
\centering
\setlength\aboverulesep{0pt}\setlength\belowrulesep{0pt}
\resizebox{1\linewidth}{!}{%
\begin{tabular}{| l || c | c | c | c | c |} 
\toprule[1.5pt]
\multicolumn{1}{| c ||}{\cellcolor[HTML]{EFEFEF}Name}
& \multicolumn{1}{c|}{\cellcolor[HTML]{EFEFEF}Country}
& \multicolumn{1}{c|}{\cellcolor[HTML]{EFEFEF}Clinical Source}
& \multicolumn{1}{c|}{\cellcolor[HTML]{EFEFEF}Clinical Context}
& \multicolumn{1}{c|}{\cellcolor[HTML]{EFEFEF}Samples}
& \multicolumn{1}{c|}{\cellcolor[HTML]{EFEFEF}Tokens (M)}
\\
\midrule
ACI-BENCH \citep{yimAcibenchNovelAmbient2023} & US & Clinical Notes & Not Reported & 207 & 0.1 \\
ADE Corpus \citep{gurulingappaDevelopmentBenchmarkCorpus2012} & Several & Clinical Notes & Not Reported & 20,896 & 0.5 \\
Brain MRI Stroke \citep{kimNaturalLanguageProcessing2019} & Korea & Radiology Reports & Neurology & 2,603 & 0.2 \\
CheXpert Plus \citep{chambonCheXpertAugmentingLarge2024} & US & Radiology Reports & Pulmonology & 223,460 & 60.6 \\
CHIFIR \citep{rozovaDetectingEvidenceInvasive2023} & Australia & Pathology Reports & Hematology / Oncology & 283 & 0.1 \\
CORAL \citep{sushilCORALExpertCuratedMedical2024} & US & Progress Notes & Hematology / Oncology & 240 & 0.7 \\
Eye Gaze CXR \citep{karargyrisCreationValidationChest2021} & US & Radiology Reports & Pulmonology & 892 & 0.03 \\
Gout Chief Complaints \citep{osborneGoutEmergencyDepartment} & US & Chief Complaint & Internal Medicine & 8,429 & 0.2 \\
ID-68 \citep{anaziExpandingGeneticHeterogeneity2017} & UK & Clinical Notes & Psychology & 78 & 0.02 \\
Inspect \citep{huangINSPECTMultimodalDataset2023} & US & Radiology Reports & Pulmonology & 22,259 & 2.8 \\
MedNLI \citep{romanovLessonsNaturalLanguage2018} & US & Clinical Notes & Internal Medicine & 14,047 & 0.5 \\
MedQA \citep{jinWhatDiseaseDoes2020} & US & National Medical Board Examination & Not Reported & 14,366 & 2.0 \\
MIMIC-III \citep{johnsonMIMICIIIFreelyAccessible2016} & US & Clinical Notes & Internal Medicine & 2,021,411 & 1,047.7 \\
MIMIC-IV Note \citep{johnsonMIMICIVFreelyAccessible2023} & US & Clinical Notes & Internal Medicine & 2,631,243 & 1,765.7 \\
MTSamples \citep{MedicalTranscriptions} & Not Reported & Clinical Notes & Internal Medicine & 2,358 & 1.7 \\
Negex \citep{chapmanExtendingNegExLexicon2013} & US & Discharge Summaries & Not Reported & 2,056 & 0.1 \\
PriMock57 \citep{korfiatisPriMock57DatasetPrimary2022} & UK & Simulated Patient Care & Internal Medicine & 57 & 0.01 \\
Q-Pain \citep{logeQPainQuestionAnswering2021} & US & Clinical Vignettes & Palliative Care & 51 & 0.01 \\
REFLACX \citep{bigolinlanfrediREFLACXDatasetReports2022} & US & Radiology Reports & Pulmonology & 2,543 & 0.1 \\
Simulated Resp. Interviews \citep{fareezDatasetSimulatedPatientphysician2022} & Canada & Simulated Patient Care & Pulmonology & 272 & 0.6 \\
\bottomrule[1.5pt]
\end{tabular}
}
\caption{Clinical datasets used for Pre-Training, with token counts in millions.}
\label{tab:clinical-datasets}
\end{table*}

\subsection{Pre-Training}

\subsubsection{Biomedical Datasets}

We use PubMed abstracts and PMC full-text articles as the sources for biomedical text. Using their respective APIs, we gather a total of 50.7B tokens of biomedical text for pretraining.

\subsubsection{Clinical Datasets}

Since the release of Clinical BERT, Clinical Longformer and Clinical BigBird, the landscape of available clinical text has evolved significantly with the introduction of MIMIC-IV~\citep{johnsonMIMICIVFreelyAccessible2023}, which increased the volume of clinical tokens available for model training. We incorporate MIMIC-IV into our clinical pretraining corpus and further diversify it by collecting additional datasets from a variety of institutions, countries, and clinical contexts. This addresses a key limitation of existing clinical encoders, which tend to rely on a single institution as their source of clinical text. The broader and more diverse dataset foundation should improve generalization and better capture variability in care practices across institutions, as well as stylistic and structural differences among individual clinicians.
 
Following a recent review by~\citeauthor{lit_review_clin_datasets}, we curate a set of 20 publicly available, non-synthetic clinical text datasets. Because ModernBERT is trained exclusively on English-language documents, we limit our selection to datasets containing English text. Further details, including the datasets' countries of origin, clinical sources and clinical contexts are provided in \autoref{tab:clinical-datasets}. The clinical datasets together amount to 2.8B tokens.

One known limitation of current clinical encoders, as noted by~\citet{alsentzerPubliclyAvailableClinical2019}, is their poor performance on de-identification tasks, which is largely due to the use of placeholder tokens for PHI in MIMIC-III. To mitigate this, we include datasets such as CheXpert Plus~\citep{chambonCheXpertAugmentingLarge2024} which replace PHI with synthetic surrogates rather than masking tokens.

Finally, while MIMIC-III and MIMIC-IV have a small degree of overlap, we do not anticipate this to adversely affect pretraining. We use a high masked language modeling (MLM) probability of 30\%, following the implementation in~\citet{modernbert}, and train for 6 epochs over the clinical dataset during the entire pretraining strategy. For comparison, RoBERTa~\citep{liuRoBERTaRobustlyOptimized2019} was trained for 40 epochs using a 15\% MLM probability. We therefore do not expect the training to be affected by this redundancy, and do not seek to exclude overlapping samples.

\subsubsection{Training Schedule}

ModernBERT was trained using a Warmup-Stable-Decay (WSD) learning rate scheduler~\citep{zhaiScalingVisionTransformers2022, huMiniCPMUnveilingPotential2024}, which consists of three stages: a warmup stage that gradually increases the learning rate, a stable stage that maintains it at a constant peak to accelerate training, and a decay stage that reduces the learning rate to ensure convergence on the training data. This schedule enables continued pretraining on new data from any checkpoint within the stable stage without cold restart issues~\citep{hägele2024scalinglawscomputeoptimaltraining, ashWarmStartingNeuralNetwork2020} and allows reuse of the stable-stage learning rate without needing a new warmup~\citep{wen2024understandingwarmupstabledecaylearningrates}, ultimately followed by a decay stage to achieve the best performance on the new target domain.

This training paradigm is particularly well suited for domain adaptation via continued pretraining: resuming from a checkpoint within the stable stage allows efficient training on new domain-specific data without reintroducing warmup instability, while the decay phase facilitates targeted specialization to the downstream domain. We apply this strategy in the development of \modelname through a two-phase training procedure inspired by the methodology behind Clinical BERT. In that case, the model was initialized from BioBERT which was trained on PubMed and PMC, and adapted to clinical language using MIMIC-III. Similarly, we aim to balance broad biomedical knowledge with clinical specificity by first resuming the stable phase training from ModernBERT’s pre-decay checkpoint on a mixture of biomedical and clinical corpora, followed by a decay phase to specialize on clinical data.

\paragraph{Phase 1}The first phase involves joint pretraining on biomedical and clinical data, specifically PubMed, PMC and the 20 curated clinical datasets, for a total of 160.5B tokens. We include clinical data in this phase to mitigate catastrophic forgetting of biomedical knowledge during the second training phase~\citep{ibrahimSimpleScalableStrategies2024}.

Accordingly, we initialize \modelname from the final stable-stage checkpoints of the ModernBERT base and large models, continuing training with their respective learning rates, 3e-4 for the base model and 5e-5 for the large model. To preserve consistency with the original training setup, we also adopt the same batch sizes: 72 for the base model and 77 for the large model.

\paragraph{Phase 2}In the second phase, we further refine the model on the 20 clinical datasets alone. For this stage, we lower the MLM probability to 15\%, following findings from~\citeauthor{anknerDynamicMaskingRate2024}, and confirm through our own experiments that this setting improves downstream performance compared to a 30\% masking ratio.

\modelname is trained for three epochs on the clinical data during this phase. We experiment with several learning rate schedules and find that the base model achieves the best downstream results with a $1-$sqrt decay applied across all three epochs. For the large model, the best performance is achieved using a constant learning rate for the first two epochs followed by a $1-$sqrt decay in the final epoch.

We also trained a separate model following Phase 1, using only biomedical data for an additional 50.7B tokens with a $1-$sqrt decay learning rate schedule. However, this model consistently underperformed in our downstream evaluations compared to the variant trained on clinical data. While we do not focus on this model further in the paper, we release it publicly on Hugging Face as Bio ModernBERT \href{https://huggingface.co/thomas-sounack/Bio-ModernBERT-base}{base} and \href{https://huggingface.co/thomas-sounack/Bio-ModernBERT-large}{large}.

\subsubsection{Computing Resources}

\modelname was trained on a server equipped with 8 NVIDIA H100 SXM5 GPUs. The continued pretraining process took approximately four days for the base model and eight days for the large model, excluding time spent on data preprocessing and ablation studies. 

\subsection{Downstream Evaluation}

\subsubsection{Models}

We compare our models to popular clinical and biomedical encoders. We provide a summary of the training data used for each model along with the corresponding token counts in \autoref{encoders-training-corpora}.

\paragraph{Clinical-Longformer and Clinical-BigBird} \citep{liClinicalLongformerClinicalBigBirdTransformers2022} are the only long-context clinical encoders available with a sequence length of 4096 tokens. They are based on the Longformer~\citep{beltagyLongformerLongDocumentTransformer2020} and BigBird~\citep{zaheerBigBirdTransformers2021} architectures, respectively, and further trained on clinical notes from MIMIC-III.

\paragraph{BioBERT} \citep{biobert} is initialized on BERT-base~\citep{devlinBERTPretrainingDeep2019}, and trained on PubMed and PMC data. We use the \href{https://huggingface.co/dmis-lab/biobert-v1.1}{biobert-v1.1 model} hosted on Hugging Face.

\paragraph{BioMed-RoBERTa} \citep{gururanganDontStopPretraining2020} is based on RoBERTa-base~\citep{liuRoBERTaRobustlyOptimized2019}, and trained on scientific papers from Semantic Scholar~\citep{kinneySemanticScholarOpen2025}.

\paragraph{Clinical BERT} \citep{alsentzerPubliclyAvailableClinical2019} introduces several models. We use the \href{https://huggingface.co/emilyalsentzer/Bio_ClinicalBERT}{Bio\_ClinicalBERT model} hosted on Hugging Face, which corresponds to the version trained on top of BioBERT using clinical notes from MIMIC-III.

\paragraph{Clinical ModernBERT} was made available at the time of writing this paper in~\citet{leeClinicalModernBERTEfficient2025}. The authors fine-tuned ModernBERT base on 13B tokens of PubMed abstracts, MIMIC-IV clinical notes and pairs of ICD codes with their description. Clinical ModernBERT was trained on the post-decay version of ModernBERT using 128-token sequences, which may limit its ability to capture long-range dependencies. In contrast, we train both base and large versions with 8192-token inputs, a substantially larger and more diverse corpus, and continue pretraining from the pre-decay checkpoints - in line with the WSD schedule used by ModernBERT.\\

We also compare our models to ModernBERT-base and ModernBERT-large, to demonstrate the added value of the continued pre-training.

\subsubsection{Tasks}

We evaluate the models on five datasets, one multi-label classification task, one single-label classification task and three Named Entity Recognition (NER) tasks. We evaluate the performance of the models by fine-tuning them using five different seeds and report the median scores on the test sets. The models are trained for 10 epochs with early stopping, with a batch size of 16 and a weight decay of $1e^{-5}$. Following standard practices, we determine the learning rate for base and large models by conducting a grid search for each downstream task~\citep{biobert, alsentzerPubliclyAvailableClinical2019, liClinicalLongformerClinicalBigBirdTransformers2022}. The values used can be found in \autoref{ft-learning-rates}. The implementations of the fine-tuning and evaluation can be found in our \href{https://github.com/lindvalllab/BioClinical-ModernBERT}{GitHub repository}.

For the classification tasks, we measure the performance using the weighted F1 score. For the NER tasks, we use the F1 score provided by the seqeval framework~\citep{seqeval}.

\paragraph{Classification tasks} 

\paragraph{ChemProt} \citep{chemprot} is a dataset containing 1,820 PubMed abstracts with chemical-protein interactions. We gather the data from the BLUE benchmark code repository~\citep{peng2019transfer}. Following their implementation, the associated task is a single-label classification problem on six classes: CPR:3, CPR:4, CPR:5, CPR:6, CPR:9 and None.

\paragraph{Phenotype} \citep{moseleyCorpusDetectingHighContext2020} provides annotations for 2,270 patient notes of MIMIC-III on the presence or absence of 15 phenotypes (including None and Unsure). Each note is annotated by two expert human annotators. Since the "None" category corresponds to cases where all phenotype indicators are absent (i.e., a vector of zeros), this task is treated as a multi-label classification problem over the remaining 14 phenotypes.

\paragraph{Named Entity Recognition Tasks}

\paragraph{Change-Of-State (COS)} or Clinical Events with Change-Of-State in Chest X-Ray Sentences~\citep{klassenAnnotatingClinicalEvents} is a corpus of 1008 sentences extracted from 1,344 de-identified UW Harborview Medical Center chest X-Ray notes. The dataset provides annotations for clinical events defined with 4 entities: anatomical location, attribute of the location, possible value for the attribute, and change-of-state for the attribute. These annotations were generated by two human experts. The associated task corresponds to a token-level multi-label classification for these four clinical events.

\paragraph{Social History} or Lifestyle and Environmental Factors in Social History Sections~\citep{yetisgenAutomaticIdentificationSubstance2017} is a corpus of 364 clinical notes' social history sections from MTSamples. Each of these sections is annotated by a human annotator for three types of substance abuse: tobacco, alcohol, and drug, with 7 entity types per event: status, type, method, amount, frequency, exposure-history and quit-history. In addition, the dataset contains annotations for the following factors: occupation, marital status, family information, residence, living situation, environmental exposure, physical activity, weight management, sexual history, infectious disease history. The associated task corresponds to a token-level multi-label classification for these social history factors.

\paragraph{De-Identified Medical Text (DEID)} \citep{neamatullahAutomatedDeidentificationFreetext2008} is a corpus of 2,434 nursing notes from the MIMIC-II database~\citep{saeedMultiparameterIntelligentMonitoring2011} annotated by three or more experts for instances of PHI. The dataset is available in two versions: one where PHI instances have been replaced by realistic surrogate data, and another where PHI instances have been replaced by corresponding tags. Using the tags, we categorize the PHI instances as one of the following PHI: age, date, location, name, phone number or other (the corresponding script can be found in our repository). The models are then presented with the text that contains realistic surrogate data. The associated task corresponds to a token-level multi-label classification for these types of PHI.

We preprocess each NER dataset to frame the corresponding task in the BIO format~\citep{ramshawTextChunkingUsing1995}. The details of the implementation can be found in our \href{https://github.com/lindvalllab/BioClinical-ModernBERT}{GitHub repository}. Statistics for each dataset can be found in \autoref{downstream-datasets-stats}.

\begin{table*}[hb!]
\centering
\renewcommand{\arraystretch}{1.2}
\resizebox{\linewidth}{!}{%
\begin{tabular}{c l c cc ccc}
\toprule
\noalign{\vskip 0.4em}
\multirow{2}{*}{} & \multirow[b]{2}{*}{\textbf{Model}} & \multirow[b]{2}{*}{\textit{Context length}} & \multicolumn{2}{c}{\textbf{Classification}} & \multicolumn{3}{c}{\textbf{Named Entity Recognition}}\\
\cmidrule(lr){4-5} \cmidrule(lr){6-8}
& & & ChemProt & Phenotype & COS & Social History & DEID \\
\noalign{\vskip 0.3em}
\hline
\noalign{\vskip 0.4em}
\multirow{8}{*}{\rotatebox[origin=c]{90}{\textbf{Base}}} 
& BioBERT \citep{biobert} & \textit{512} & 89.5 & 26.6 & 94.9 & 55.8 & 74.3 \\
& Clinical BERT \citep{alsentzerPubliclyAvailableClinical2019} & \textit{512} & 88.3 & 25.8 & 95.0 & 55.2 & 74.2 \\
& BioMed-RoBERTa \citep{gururanganDontStopPretraining2020} & \textit{512} & 89.0 & 36.8 & 94.9 & 55.2 & 81.1 \\
& Clinical-BigBird \citep{liClinicalLongformerClinicalBigBirdTransformers2022} & \textit{4096} & 87.4 & 26.5 & 94.0 & 53.3 & 71.2 \\
& Clinical-Longformer \citep{liClinicalLongformerClinicalBigBirdTransformers2022} & \textit{4096} & 74.2 & 46.4 & \textbf{\underline{95.2}} & 56.8 & 82.3 \\
& Clinical ModernBERT \citep{leeClinicalModernBERTEfficient2025} & \textit{8192} & 86.9 & 54.9 & 93.7 & 53.8 & 44.4 \\
& ModernBERT - base \citep{modernbert} & \textit{8192} & 89.5 & 48.4 & 94.0 & 53.1 & 78.3 \\
& \cellcolor{gray!15}\modelname\ - base (ours) & \cellcolor{gray!15}\textit{8192} & \cellcolor{gray!15}\underline{89.9} & \cellcolor{gray!15}\underline{58.1} & \cellcolor{gray!15}95.1 & \cellcolor{gray!15}\textbf{\underline{58.5}} & \cellcolor{gray!15}\underline{82.7} \\
\noalign{\vskip 0.4em}
\hline
\noalign{\vskip 0.4em}
\multirow{2}{*}{\rotatebox[origin=c]{90}{\textbf{Large}}}
& ModernBERT - large \citep{modernbert} & \textit{8192} & 90.2 & 58.3 & 94.4 & 54.8 & 82.1 \\
& \cellcolor{gray!15}\modelname\ - large (ours) & \cellcolor{gray!15}\textit{8192} & \cellcolor{gray!15}\textbf{90.8} & \cellcolor{gray!15}\textbf{60.8} & \cellcolor{gray!15}95.1 & \cellcolor{gray!15}57.1 & \cellcolor{gray!15}\textbf{83.8} \\
\end{tabular}%
}
\caption{Median performance of models on Classification and NER tasks over 5 seeds. The best score overall is in bold and the best score for base models is underlined.}
\label{tab:model-performance}
\end{table*}

\subsubsection{Inference Speed}

We adopt the inference speed evaluation methodology described in~\citet{modernbert}. To simulate real-world usage, we construct six synthetic datasets, each containing 8,192 documents. The first three datasets consist of fixed-length samples with 512, 4,096, and 8,192 tokens per document, corresponding to the three context lengths supported by the models under evaluation. To assess the impact of unpadding, we additionally generate three variable-length datasets in which token counts per sample follow a normal distribution centered at half the maximum sequence length: 256, 2,048, and 4,096 tokens, respectively. Inference speed is averaged over 10 runs for each dataset.

\subsubsection{Computing Resources}

Fine-tuning on downstream benchmark datasets was performed on a single H100 PCIe GPU, while inference speed evaluations were conducted on a machine with a single A100 40GB GPU.
\section{Results}

\subsection{Performance}

\autoref{tab:model-performance} presents the performance of various biomedical and clinical encoders across the tasks described in the previous section. On four of the five tasks, \modelname base and large outperform the other models.

For classification tasks, \modelname large achieves state-of-the-art results with an F1 score of 90.8\% on ChemProt and 60.8\% on Phenotype. The base model also outperforms all other base models, achieving 89.9\% on ChemProt and 58.1\% on Phenotype. In named entity recognition, the base model achieves state-of-the-art performance on Social History and outperforms other base models on DEID. The large version achieves state-of-the-art results on DEID, and both \modelname models remain competitive on COS with 95.1\%.

To evaluate biomedical knowledge retention after clinical specialization, we test the Phase 1 checkpoint on ChemProt, a dataset of PubMed abstracts that serves as a proxy for biomedical domain knowledge. The base model achieves an F1 score of 90.2\%, while the large model scores 90.5\%. After Phase 2, the base model scores 89.9\% and the large model improves to 90.8\%, surpassing its Phase 1 score. For comparison, BioBERT scores 89.5\% on ChemProt, while Clinical BERT scores 88.3\%.

\subsection{Inference Speed}

\begin{table*}[hb!]
\centering
\renewcommand{\arraystretch}{1.2}
\resizebox{\linewidth}{!}{%
\begin{tabular}{c l cc cc cc}
\toprule
\noalign{\vskip 0.4em}
\multirow{2}{*}{} & \multirow[b]{2}{*}{\textbf{Model}} & \multicolumn{2}{c}{\textbf{Short}} & \multicolumn{2}{c}{\textbf{Medium}} & \multicolumn{2}{c}{\textbf{Long}} \\
\cmidrule(lr){3-4} \cmidrule(lr){5-6} \cmidrule(lr){7-8}
& & Fixed & Variable & Fixed & Variable & Fixed & Variable \\
\noalign{\vskip 0.3em}
\hline
\noalign{\vskip 0.4em}
\multirow{8}{*}{\rotatebox[origin=c]{90}{\textbf{Base}}} 
& BioBERT \citep{biobert} & \textbf{98.1} & 49.3 & -- & -- & -- & -- \\
& Clinical BERT \citep{alsentzerPubliclyAvailableClinical2019} & \textbf{98.1} & 49.3 & -- & -- & -- & -- \\
& BioMed-RoBERTa \citep{gururanganDontStopPretraining2020} & 97.4 & 48.6 & -- & -- & -- & -- \\
& Clinical-BigBird \citep{liClinicalLongformerClinicalBigBirdTransformers2022} & 71.5 & 35.7 & 50.1 & 25.0 & -- & -- \\
& Clinical-Longformer \citep{liClinicalLongformerClinicalBigBirdTransformers2022} & 55.5 & 27.8 & 53.2 & 26.6 & -- & -- \\
& Clinical ModernBERT \citep{leeClinicalModernBERTEfficient2025} & 86.2 & 43.1 & 44.4 & 19.8 & 28.8 & 12.3 \\
& ModernBERT - base \citep{modernbert} & 76.2 & 75.0 & \textbf{73.8} & 74.9 & \textbf{71.1} & \textbf{73.7} \\
& \cellcolor{gray!15}\modelname\ - base (ours) & \cellcolor{gray!15}76.2 & \cellcolor{gray!15}\textbf{75.1} & \cellcolor{gray!15}\textbf{73.8} & \cellcolor{gray!15}\textbf{75.0} & \cellcolor{gray!15}\textbf{71.1} & \cellcolor{gray!15}\textbf{73.7} \\
\noalign{\vskip 0.4em}
\hline
\noalign{\vskip 0.4em}
\multirow{2}{*}{\rotatebox[origin=c]{90}{\textbf{Large}}}
& ModernBERT - large \citep{modernbert} & 25.8 & 26.5 & 25.3 & 25.6 & 24.8 & 25.2 \\
& \cellcolor{gray!15}\modelname\ - large (ours) & \cellcolor{gray!15}25.8 & \cellcolor{gray!15}26.5 & \cellcolor{gray!15}25.3 & \cellcolor{gray!15}25.6 & \cellcolor{gray!15}24.8 & \cellcolor{gray!15}25.2 \\
\end{tabular}%
}
\caption{Inference speed in thousands of tokens per second (kTok/s), averaged over 10 runs. The best score for each category is in bold. Dashes indicate unsupported configurations.}
\label{tab:model-speed}
\end{table*}

Results of the inference speed testing are presented in \autoref{tab:model-speed}. Among the base models evaluated, ModernBERT demonstrated the highest overall efficiency. While short-context encoders like BERT and RoBERTa perform well on datasets with fixed input sizes of 512 tokens, achieving 98.1 and 97.4 thousand tokens per second respectively, ModernBERT still delivers competitive throughput at 76.2 kTok/s and outperforms other long-context models such as BigBird (71.5 kTok/s) and Longformer (55.5 kTok/s) on the same task. Across all other dataset types, ModernBERT consistently provides the best performance, maintaining a relatively stable processing speed across all input lengths: 76.2 kTok/s at 512 tokens, 73.8 kTok/s at 4096 tokens, and 71.1 kTok/s at 8192 tokens for fixed datasets. In contrast, BigBird's throughput drops significantly from 71.5 to 50.1 kTok/s on fixed datasets as input length increases. Moreover, ModernBERT exhibits significantly stronger performance on datasets with variable-length sequences, reaching 75.1 kTok/s for the 512-token dataset and 73.7 kTok/s for the 8192-token dataset.

The model proposed by~\citeauthor{leeClinicalModernBERTEfficient2025}, although initialized on ModernBERT base, exhibited behaviour that was inconsistent with the ModernBERT architecture in our inference speed tests. For datasets containing sequences of fixed length, it is faster on 512 tokens inputs. However, the throughput significantly degrades as the inputs get longer, offering worse performance than ModernBERT on medium and long sequences. For datasets containing sequences of variable length, it performs worse than for fixed length datasets, suggesting that unpadding is not being used. Upon further inspection of the model's configuration file, it appears that it is following a BERT architecture with a context length of 8192 tokens, which explains the efficiency results we observed.
\section{Discussion}

\modelname base and large demonstrate strong and consistent performance in both classification and named entity recognition tasks, outperforming existing clinical and biomedical encoders. The significant performance gains observed on Phenotype can be attributed to both the model's improved adaptation to clinical language and its ability to leverage the significantly longer input sequences present in this dataset (see  \autoref{downstream-datasets-stats}).

Notably, our results show that \modelname retained substantial biomedical knowledge even after Phase 2 specialization on clinical data. This is evidenced by the small change in ChemProt performance between Phase 1 and Phase 2: a slight decline of 0.3 points for the base model and an improvement of 0.3 points for the large model. These findings suggest that incorporating clinical data during the first stage of pretraining helps prevent catastrophic forgetting, allowing \modelname to preserve biomedical knowledge throughout subsequent clinical adaptation. In contrast, Clinical BERT scored 1.2 points lower on ChemProt than its biomedical-only predecessor, BioBERT, indicating some degradation in biomedical performance following its second-phase clinical training.

We also found evidence that our approach addresses a challenge previously noted by the authors of Clinical BERT: the reduced performance of clinical embedding models on de-identification tasks, often attributed to the use of generic PHI masking tokens in datasets like MIMIC. Our results on the DEID task support the hypothesis that training on a diverse set of clinical datasets, with a range of PHI handling strategies, enhances the model’s ability to recognize and generalize over protected health information. In particular, some datasets in our corpus replace PHI with synthetic but realistic surrogates rather than masking tokens, which may help preserve semantic structure. This variation in PHI representation appears to improve the model’s comprehension of PHI during fine-tuning, leading to stronger downstream performance on de-identification tasks.

In addition to strong task performance, our results show that \modelname is the only clinical encoder that consistently maintains high computational efficiency across a wide range of sequence lengths and input distributions. While short-context encoders exhibit faster runtimes on datasets with short, fixed-length inputs, which is partly due to their smaller parameter counts (110M for BERT base and 125M for RoBERTa base, compared to ModernBERT base's 150M), \modelname remains competitive in such scenarios and demonstrates clear advantages as input lengths increase. This is enabled by its alternating attention mechanism, in which two-thirds of attention layers are restricted to local sliding windows rather than full-sequence global attention, substantially reducing complexity. 

Efficiency on variable-length inputs is further enhanced by ModernBERT’s unpadding mechanism, which dynamically excludes padding tokens during inference. Unlike standard transformer implementations that compute attention over uniformly padded sequences, unpadding ensures that computation is performed only on actual content tokens by ignoring padding tokens, reducing memory usage and speeding up inference. This property is especially valuable in clinical NLP where input lengths vary significantly across note types and patient encounters, leading to considerable inefficiencies when using static padding.
            
\paragraph{Limitations and Future Work}
Our work has several limitations. First, while we sought to provide \modelname with good clinical generalization capabilities by using a variety of clinical datasets, MIMIC-III and MIMIC-IV still constituted over 95\% of the clinical training corpus; this limits the true clinical data diversity we were able to achieve. Second, as ModernBERT was trained on the English language only, we made the decision to train \modelname exclusively with English clinical notes. Our work is therefore not directly applicable to other languages. Finally, With respect to benchmarking datasets, i2b2 datasets~\citep{uzunerEvaluatingStateoftheartAutomatic2007, uzuner2010I2b2VA2011, sunAnnotatingTemporalInformation2013, stubbsAutomatedSystemsDeidentification2015, stubbsAnnotatingLongitudinalClinical2015} are typically used to benchmark clinical encoders. While we had planned to benchmark \modelname on these to be consistent with the literature, the datasets were unavailable at the time of this project. We plan to release benchmarks for \modelname on these datasets as soon as they are made available again.
\section{Conclusion}

In this work, we introduce \modelname, a long-context encoder designed for clinical and biomedical text. Building on the foundation of ModernBERT and leveraging its training schedule, we employ a two-phase continued pre-training strategy that combines large-scale biomedical corpora with a diverse and extensive collection of publicly available clinical data, spanning 20 datasets from multiple countries, institutions, and clinical settings. To our knowledge, this constitutes the largest biomedical and clinical pretraining corpus used for training an encoder to date.

We release both base and large versions of \modelname, enabling researchers and practitioners to choose between computational efficiency and performance depending on their use case. Our pre-training design addresses a key limitation of previous clinical models, which rely almost exclusively on MIMIC-III, and results in improved generalization across clinical NLP tasks. By integrating biomedical and clinical data early in training, and leveraging long-context modeling capacity, \modelname achieves state-of-the-art performance across a broad set of clinical benchmarks.

Moreover, \modelname supports long-context inputs up to 8,192 tokens. This is particularly critical in medical NLP, where important information is often distributed across long sequences, and clinical reasoning depends on capturing long-range dependencies and document-level context. Despite this extended context, the model remains highly efficient, offering fast inference even at long input lengths.

We release our code, models, and training checkpoints to support further research.
\section{Acknowledgments}

We gratefully acknowledge the Stanford Center for Artificial Intelligence in Medicine and Imaging, as well as PhysioNet~\citep{goldbergerPhysioBankPhysioToolkitPhysioNet2000} for providing access to several of the clinical datasets utilized in this study. We also thank Orion Weller for his valuable assistance in troubleshooting issues related to the tokenization process.

\section{Code availability statement}

We provide the \modelname models\\
(\href{https://huggingface.co/thomas-sounack/BioClinical-ModernBERT-base}{base}, \href{https://huggingface.co/thomas-sounack/BioClinical-ModernBERT-large}{large})  and  \href{https://huggingface.co/thomas-sounack/BioClinical-ModernBERT-checkpoints}{training checkpoints} on Hugging Face, and the \href{https://github.com/lindvalllab/BioClinical-ModernBERT}{source code} for reproducibility on GitHub.

\bibliography{anthology,custom}
\bibliographystyle{acl_natbib}

\clearpage

\onecolumn 

\appendix

\section{Encoders training corpora}\label{encoders-training-corpora}

\autoref{tab:training-data-tokens} summarizes the training data composition for each model in terms of clinical, biomedical, and other sources, along with token counts where available. This breakdown highlights the diversity and scale of data used to train \modelname (ours) compared to prior models.\\

\begin{table}[h]
\centering
\resizebox{\linewidth}{!}{%
\begin{tabular}{l
                l r  
                l r  
                l r  
                r    
                }
\toprule
\textbf{Model}
& \multicolumn{2}{c}{\textbf{Clinical}} 
& \multicolumn{2}{c}{\textbf{Biomedical}} 
& \multicolumn{2}{c}{\textbf{Other}} 
& \textbf{Total} \\
\cmidrule(lr){2-3} \cmidrule(lr){4-5} \cmidrule(lr){6-7}
& Source              & Tokens 
& Source              & Tokens 
& Source                           & Tokens 
& (B tokens) \\
\midrule
BioBERT              
& —                   & —      
& PubMed + PMC        & 18.0   
& Wikipedia + BookCorpus      & 3.3    
& 21.3        \\

Clinical BERT        
& MIMIC-III           & 1.0    
& PubMed + PMC        & 18.0   
& Wikipedia + BookCorpus      & 3.3    
& 22.3        \\

BioMed-RoBERTa       
& —                   & —      
& S2ORC papers        & 7.6    
& —                       & —      
& 7.6         \\

Clinical-Longformer  
& MIMIC-III           & 1.0    
& —                   & —      
& —                       & —      
& 1.0         \\

Clinical-BigBird     
& MIMIC-III           & 1.0    
& —                   & —      
& —                       & —      
& 1.0         \\

Clinical ModernBERT  
& MIMIC-IV            & 1.7    
& PubMed              & (N/A)  
& Medical codes \& description    & (N/A)  
& 13.0        \\

\cellcolor{gray!15}\modelname (ours)
& \cellcolor{gray!15}20 clinical datasets& \cellcolor{gray!15}\textbf{2.8}   
& \cellcolor{gray!15}PubMed + PMC        & \cellcolor{gray!15}\textbf{50.7}   
& \cellcolor{gray!15}—                       & \cellcolor{gray!15}—      
& \cellcolor{gray!15}\textbf{53.5} \\

\bottomrule
\end{tabular}%
}
\caption{Training data composition (in billions of tokens) by data type.}
\label{tab:training-data-tokens}
\end{table}

\section{Fine-tuning learning rates}\label{ft-learning-rates}

\autoref{tab:task-learning-rates} reports the selected learning rates for fine-tuning the encoders on each downstream task. Separate values are shown for the base and large model variants.\\

\begin{table}[h]
\centering
\resizebox{0.3\linewidth}{!}{%
\begin{tabular}{lcc}
\toprule
\textbf{Task} & \textbf{Base} & \textbf{Large} \\
\midrule
ChemProt         & 5e-5 & 2e-5 \\
Phenotype        & 8e-5 & 5e-5 \\
COS              & 1e-4 & 1.5e-4 \\
Social History   & 1.5e-4 & 2e-4 \\
DEID             & 7e-5 & 7e-5 \\
\bottomrule
\end{tabular}%
}
\caption{Selected learning rates for each task and model size.}
\label{tab:task-learning-rates}
\end{table}

\section{Downstream evaluation datasets description}\label{downstream-datasets-stats}

\autoref{tab:dataset-metrics} summarizes the dataset statistics, including the number of classes for classification datasets or BIO-tagged entity types for NER datasets, the number of examples per split, and the average number of tokens per example computed using the ModernBERT tokenizer.

\begin{table*}[ht!]
\centering
\renewcommand{\arraystretch}{1.2}
\resizebox{0.8\linewidth}{!}{%
\begin{tabular}{c l c c c c c c c}
\toprule
\noalign{\vskip 0.4em}
\multirow{2}{*}{} & \multirow[b]{2}{*}{\textbf{Dataset}} & \multirow[b]{2}{*}{\textbf{\# Classes / Entities}} & \multicolumn{3}{c}{\textbf{Samples per Split}} & \multicolumn{3}{c}{\textbf{Avg. \# Tokens}} \\
\cmidrule(lr){4-6} \cmidrule(lr){7-9}
& & & \textbf{Train} & \textbf{Val} & \textbf{Test} & \textbf{Train} & \textbf{Val} & \textbf{Test} \\
\noalign{\vskip 0.3em}
\hline
\noalign{\vskip 0.4em}
\multirow{2}{*}{\rotatebox[origin=c]{90}{\textbf{Classif.}}}
& ChemProt & 6 & 19,460 & 11,820 & 16,943 & 68.2 & 67.7 & 75.2 \\
& Phenotype & 14 & 1,589 & 227 & 454 & 3,112.2 & 3,118.7 & 3,175.8 \\
\noalign{\vskip 0.4em}
\hline
\noalign{\vskip 0.4em}
\multirow{3}{*}{\rotatebox[origin=c]{90}{\textbf{NER}}}
& COS & 11 & 705 & 151 & 152 & 24.5 & 24.0 & 23.4 \\
& Social History & 51 & 254 & 55 & 55 & 55.9 & 57.0 & 44.9 \\
& DEID & 13 & 1,703 & 365 & 366 & 276.6 & 270.0 & 281.5 \\
\end{tabular}%
}
\caption{Dataset characteristics: number of classes or entity types, total examples per split, and average number of tokens per sample.}
\label{tab:dataset-metrics}
\end{table*}

\end{document}